# Robotic-assisted Ultrasound for Fetal Imaging: Evolution from Single-arm to Dual-arm System

Shuangyi Wang[1], James Housden[1], Yohan Noh[1], Davinder Singh[2], Anisha Singh[2], Emily Skelton[3], Jacqueline Matthew[3], Cornelius Tan[1], Junghwan Back[4], Lukas Lindenroth[4], Alberto Gomez[1], Nicolas Toussaint[1], Veronika Zimmer[1], Caroline Knight[5], Tara Fletcher[3], David Lloyd[6], John Simpson[6], Dharmintra Pasupathy[5], Hongbin Liu[4], Kaspar Althoefer[7], Joseph Hajnal[1], Reza Razavi[1], and Kawal Rhode[1]

[1] School of Biomedical Engineering & Imaging Sciences, King's College London, UK
[2] Xtronics Ltd., Gravesend, UK
[3] Women's Ultrasound, Guy's & St Thomas' NHS Foundation Trust, London, UK
[4] Department of Informatics, King's College London, UK
[5] Division of Women's Health, King's College London, Fetal Medicine Unit, Guy's & St Thomas' NHS Foundation Trust, Women's Health Academic Centre, King's Health Partners, London, UK
[6] Fetal Cardiology, Evelina Children's Hospital, London, UK
[7] Faculty of Science & Engineering, Queen Mary University of London, UK
shuangyi.wang@kcl.ac.uk

**Abstract.** The development of robotic-assisted extracorporeal ultrasound systems has a long history and a number of projects have been proposed since the 1990s focusing on different technical aspects. These aim to resolve the deficiencies of on-site manual manipulation of hand-held ultrasound probes. This paper presents the recent ongoing developments of a series of bespoke robotic systems, including both single-arm and dual-arm versions, for a project known as intelligent Fetal Imaging and Diagnosis (iFIND). After a brief review of the development history of the extracorporeal ultrasound robotic system used for fetal and abdominal examinations, the specific aim of the iFIND robots, the design evolution, the implementation details of each version, and the initial clinical feedback of the iFIND robot series are presented. Based on the preliminary testing of these newly-proposed robots on 42 volunteers, the successful and reliable working of the mechatronic systems were validated. Analysis of a participant questionnaire indicates a comfortable scanning experience for the volunteers and a good acceptance rate to being scanned by the robots.

## 1 Introduction

An extracorporeal robotic ultrasound system refers to the configuration in which the robotic system is constructed to hold and manipulate hand-held ultrasound probes for external examinations. The research interests in motorizing ultrasound systems started in the late 1990s within the European Union, North America, and Japan [1]. This was motivated by the deficiencies of the on-site manual manipulation of hand-held probes, such as difficulties of maintaining accurate probe positioning for long periods of time using human hands [2] and the requirements for experienced sonographers to be on-site [3]. Many of these robotic systems were designed in the typical master-slave configuration, whereby the master-side sonographer can be in a remote location to perform the examination and a slave-side robot driving the ultrasound probe mimics the movements of the remote sonographer. These systems were mainly designed for diagnostic purposes but a few of them were also aimed at guidance of interventional procedures or open surgeries.

The iFIND (intelligent Fetal Imaging and Diagnosis) project is a recent ongoing research project that relates to the use of robotic system to assist ultrasound examination. Started in 2014, this project aims to improve the accuracy of routine 18-20 week screening in pregnancy by developing new computer-guided ultrasound technologies that will allow screening of fetal abnormalities in an automated and uniform fashion. This was motivated by evidence that the diagnostic accuracy and sensitivity of ultrasound can be limited by technical restraints in the imaging. There is also strong evidence of major regional and hospital-specific variation in prenatal detection rates of major anomalies [4, 5]. Within the aim of the iFIND project, developing new ultrasound robots, which have the potential to assist and standardize the ultrasound scan, has been set as one of the objectives.

Utilization of robotic systems for fetal and abdominal examinations is one of the biggest research directions in the area of ultrasound robotics as it could include scanning of many possible anatomies and it is also one of the most easily accessible ultrasound scanning areas. One of the early robotic ultrasound systems proposed by Vilchis *et al*. [6, 7] was a unique robot aiming for abdominal examinations, known as TER. In the design, motor-driven cables were supported on the examination table. These cables translated a circular platform, upon which a mounted robotic wrist generated angular orientation. This early work has had significant influence on subsequent research, such as the work from Masuda *et al*. [8] a few years later which introduced a platform with jointed legs on a pair of rails. The leg joints along with the raising/lowering of the platform allowed 6-DOF positioning of an ultrasound probe to perform an abdominal scan.

Originally introduced by Gourdon *et al*. [9] and Arbeille *et al*. [10], a cage-like probe holder containing a robotic wrist was designed for abdominal examination. The configuration of this robot is unique as it does not include any translational axes, and was instead held in position manually at the region of scanning. The wrist incorporated three rotational axes with a unique remote-centre-of-motion mechanism, allowing a remote ultrasound expert to orient the probe locally. Supported by the European Space Agency (ESA), the projects TERESA [11] and ESTELE [12] have largely tested the proposed robot on transabdominal obstetrical and abdominal examinations for remote diagnosis. The OTELO project, developed by multiple partners within the European Union, utilized similar rotational mechanisms from the previous ESA-funded projects but added additional active translational axes to the design. The emphasis was on light weight and portability when used for general ultrasound examination [13, 14]. The research with this 6-DOF robot included a wide range of topics, such as teleoperation, kinematics, automatic control laws, and ergonomic control.

Studying the development history of extracorporeal ultrasound robots, we identified that there were very limited new bespoke systems proposed in recent years. Dur-ing this time, new rapid prototyping techniques such as 3D printing have emerged, and these have significantly changed the methods of mechanical design and manufacture. We believe that the use of 3D printing techniques offers new opportunities to design specially-shaped robot structures, which might improve the clinical acceptance and the fundamental safety of an ultrasound robot. Moreover, with the rapidly growing field of image processing and machine learning techniques, some of the fundamental difficulties of processing and interpreting ultrasound images have been addressed, which potentially changes the design requirements of an ultrasound robot, e.g. automation rather than telemedicine. Therefore, we strongly feel that it is timely to introduce a new series of ultrasound robots for the iFIND project. With several versions of robots developed and tested, this paper briefly reports the design evolution and the preliminary clinical feedback of our proposed robots. Compared with most of the previous projects on extra-corporeal ultrasound robots, the robots designed for the iFIND project are a series of robots including single-arm versions manipulating one ultrasound probe and a dual-arm version manipulating two probes simultaneously to explore novel scanning approaches. Additionally, these proposed iFIND robots do not focus on telemedicine but aim to provide a powerful research tool to explore new way of ultrasound imaging.

## 2 Design Evolution and Implementation

### 2.1 iFIND Version 1 Robot

The iFIND-v1 robot has a simple Cartesian configuration developed as a proof-of-concept prototype. The robot has seven DOFs with three orthogonal translational axes for global positioning ($J_1$, $J_2$, and $J_3$), three orthogonal rotational axes for orientation adjustments ($J_4$, $J_5$, $J_7$), and an additional translation axis ($J_6$) at the distal end of the robot to control the accurate contact of the probe with the abdominal surface.

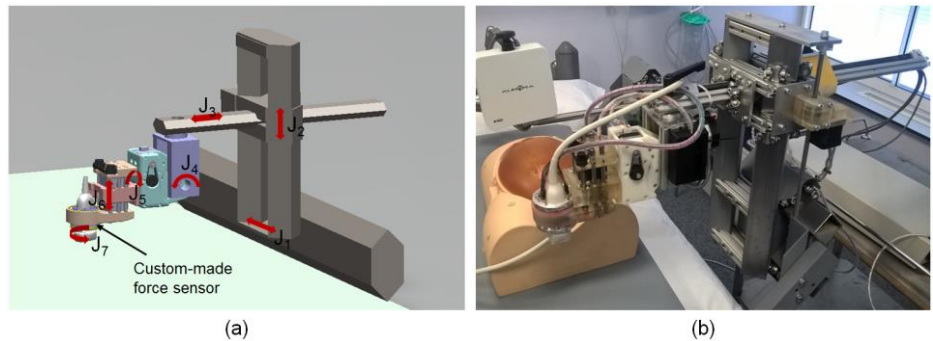

**Fig. 1.** iFIND-v1 robot: (a) schematic representation with each joint and main structures labelled and (b) final implementation of the robot shown with a fetal ultrasound phantom.

The probe holder mechanism has multiple specially shaped cavities which can include single-axis force sensors based on miniature reflective optoelectronic sensors for the measurement of vertical and side forces applied by the probe to the patient. A similar multiple-axis force sensor based on a simply-supported beam was documented in our previous research [15]. The diagram of the robot, with each joint and the main functional structures labelled, is shown in Fig. 1a along with the final implementation of the system shown in Fig. 1b. For safety management, the iFIND-v1 robot mainly relies on force control using the custom-made force sensor. With the kinematics solved, we implemented conventional robotic control methods and invited sonographers to try the system and collected feedback to guide further developments. It was generally believed that although the system provided several useful functions and can acquire ultrasound images, this industrial-looking robot working in a clinical environment with in-adequate safety features could not be clinically translated.

### 2.2 iFIND Version 2 Robot

Based on the lessons learned from the iFIND-v1 robot, we modified the design substantially by changing the shapes, configurations, mechanisms, and safety management methods of the robot, which led to the design of the iFIND-v2 robot (Fig. 2). The proposed system has a 5-DOF light-weight wrist unit [16] for holding and locally adjusting the probe ($J_4$, $J_5$, $J_6$, $J_7$, and $J_8$) and a 2-DOF two-bar arm-based set of parallel link mechanisms ($J_2$, $J_3$) with a 1-DOF rotational axis for global positioning ($J_1$). The specially designed new end-effector is lightweight and has a smaller footprint compared with the end-effector unit for the iFIND-v1 robot.

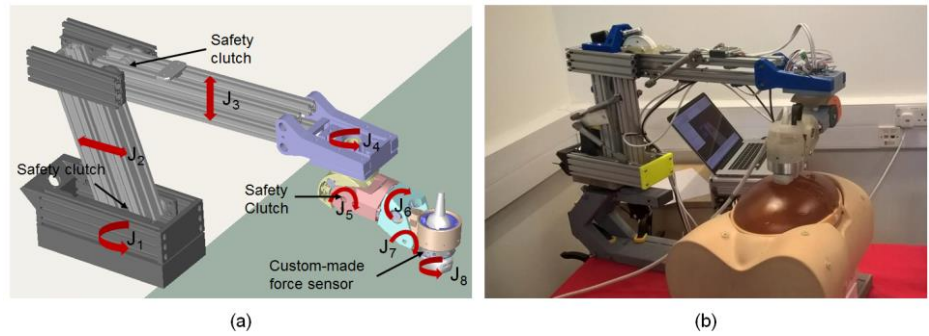

**Fig. 2.** iFIND-v2 robot: (a) schematic representation with each joint and main structures labelled and (b) final implementation of the robot shown with a fetal ultrasound phantom.

As a result of this design, the total weight of the end-effector unit is less than 2 kg and the length of the end-effector unit is about 25 cm. In terms of functionality of the joints in the new end-effector unit, J4 can rotate the following structures 360 degrees to allow the US probe to point towards different sides of the scanning area, such as the top, bottom, and sides of the abdomen. J5 is used to tilt down the probe to align with the surface of the scanning area. The last three orthogonal revolute joints ($J_6$, $J_7$, and $J_8$) are used to control the tilting and axial rotation of the probe, allowing fine adjustments of the probe in a local area. In addition to employing a similar force sen-sor to that used in the iFIND-v1 robot, the mechanical safety of the iFIND-v2 robot was emphasized with clutch mechanisms incorporated into three joints to limit the allowable force applied to the patient. These would disengage the following links from the joint driven gears when the load exceeds a pre-set threshold [16]. Additionally, gas springs were implemented to lift the robot off the patient if the clutch at the back of the robot arm is triggered.

### 2.3 iFIND Version 3 Robot

The dual-probe system has been developed directly from our experience with the iFIND-v2 single-arm robot. Several design iterations have been considered with our robotics team, clinical team, and image analysis team working together to determine a suitable design. More consideration was given to the placement of the robot arms over the patient, and how this would affect clinical and patient acceptability, as well as the working space, safety and reliability of the robot. It was agreed that a side-mounted gantry system over the patient, with the two arms attached to the gantry coming in from the side, would be the design goal for the iFIND-v3 dual-probe robot. Based on our experience testing the iFIND-v2 single-arm robot, several changes were made to the mechanical design of some joints. These joints are now made from harder-wearing materials with improved mechanisms, and all safety-critical joints now include mechanical clutches to prevent excessive force being applied.

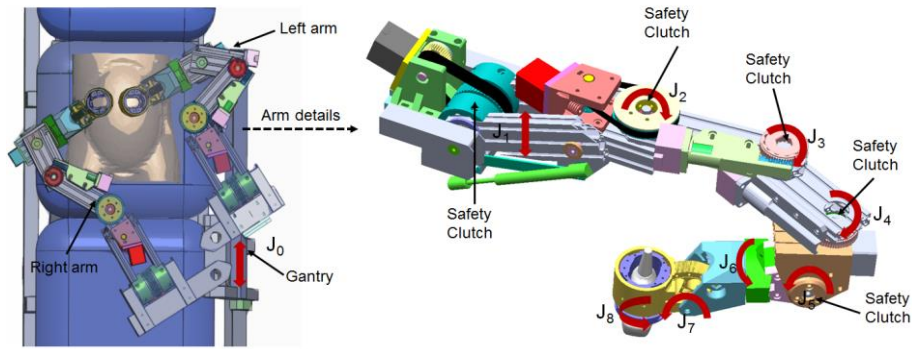

**Fig. 3.** iFIND-v3 robot: schematic representation with the general dual arm configuration shown (left) and the joint details for one arm shown (right).

Some joints have been given a larger movement range, and in particular the final end-effector is able to tilt downwards to almost vertical while keeping the probe in contact with the abdomen. This was essential for allowing the flexibility to place the two probes close together and maneuver them through a continuous sweep without the arms colliding. As illustrated in Fig. 3, the final design has 17 DOFs with two arms holding and controlling two ultrasound probes. These include one translational DOF for the gantry ($J_0$), three rotational DOFs ($J_1$, $J_2$, and $J_3$) for each of the arms, and five rotational DOFs ($J_4$, $J_5$, $J_6$, $J_7$, and $J_8$) for each of the end wrist units. The redundant DOFs in the system were designed to allow the two ultrasound probes positioned and orientated flexibly while at the same time not colliding into each other. Compared with the iFIND-v2 robot, each joint was designed to have the capability for housing a homing sensor, which allows the robot to be easily reset to its starting position. This also ensures more consistent positioning accuracy, because the starting position will be known by the control software with greater precision. Additionally, the iFIND-v3 robot is implemented on a trolley system, allowing easy transportation of the device. The final implementation of the iFIND-v3 robot is shown in Fig. 4.

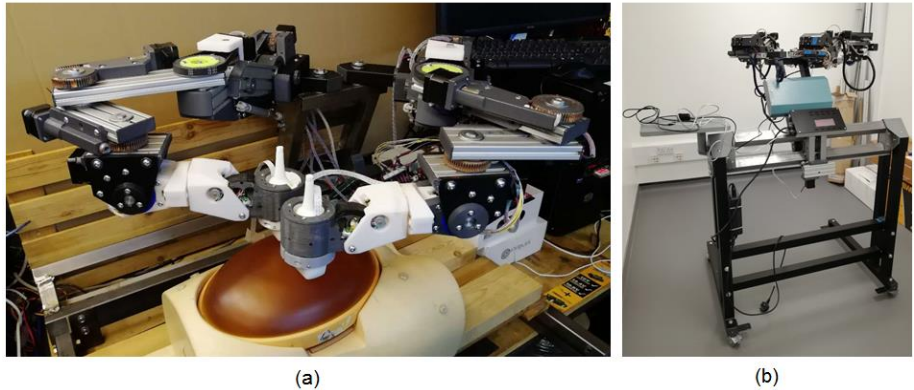

**Fig. 4.** Final implementation of the iFIND-v3 robot: (a) perspective showing the dual-arm configuration with a fetal phantom shown and (b) the trolley system for holding the arms.

## 3  Preliminary Healthy Volunteer Study

Live tests of the robots with the participation of sonographers, engineers, and most importantly volunteers greatly contributes to the further development of the systems while we are still in the design phase and able to change the configuration of the robot. After successfully and adequately testing the iFIND-v2 and iFIND-v3 robots on a fetal phantom, we applied for and obtained ethical approval to test our robots on non-pregnant volunteers for general abdominal scans. Approval was given by the King's College London local ethics committee (study title: Investigating Robotic Abdominal Ultrasound Imaging, Study reference: HR-17/18-5412). Through this study, we have successfully performed a large number of live tests. The volunteer tests started with the use of the iFIND-v2 single-probe robot, scanning 20 volunteers, and then transitioned to the iFIND-v3 multi-probe robot for testing more advanced features. So far, 22 volunteers have been scanned using the iFIND-v3 robot. The initial technical aim of the volunteer study was to test the reliability of the mechatronic system of the robots, verify the safety management methods, and experiment with potential control and image acquisition schemes. Moreover, the weekly-scheduled volunteer study intends to offer the sonographers and the engineers an opportunity to work as a team to build confidence in using the robot in a realistic scenario and overcome the psychological anxiety of the use of robotic technology in medicine. Most importantly, the volunteer study aimed to collect volunteer's feedback on the experience of being scanned by the robots, which is fedback to our design loop and influences the technical direction of the project.

Volunteer tests using the iFIND-v2 and iFIND-v3 robots are shown in Fig. 5. For the setup, the robotic system was located at the left side of the bed controlled and monitored by the engineer while the sonographer controlled the ultrasound machine on the right side of the bed. For the iFIND-v3 robot, some of the tests involved an imaging workstation to process and display images from both probes. The workstation was located at the head end of the bed where both the sonographers and engineers could observe the two images simultaneously.

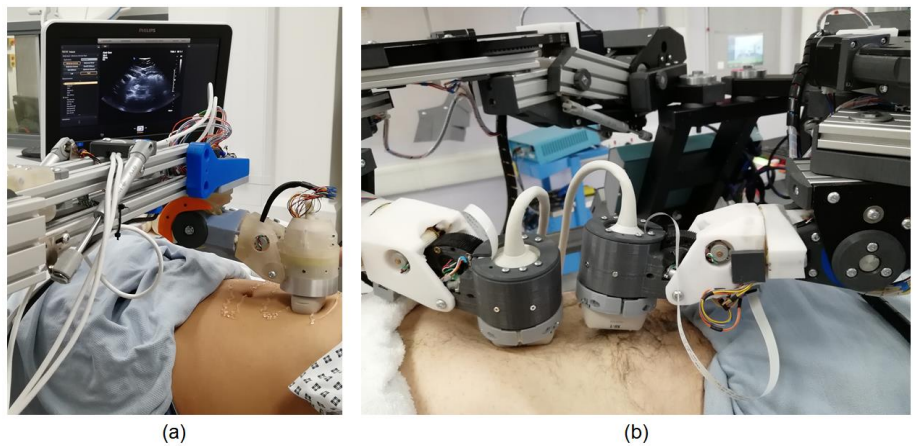

**Fig. 5.** Volunteer tests performed using (a) the iFIND-v2 robot and (b) the iFIND-v3 robot.

The most fundamental test for both robots required the sonographers to give verbal instructions to the engineers, who then controlled the robotic software and manipulated the probe accordingly to acquire standard views for a general abdominal scan. Targets included structures such as the aorta, liver, pancreas, and kidney. This was mainly to test the general reliability of the mechatronic system and focused on collecting volunteers' feedback. Moving forward, we utilized a Kinect scanner to acquire the abdominal surface of the volunteer and imported that into the robot software. Based on the kinematics, the ability of each robot to follow the acquired surface was tested. In this mode, the target positions of the probe were provided by the Kinect scan and the robots manipulated the probes to follow the abdominal surface. For the iFIND-v2 robot, control based on the force and proximity sensors was also tested in some of the sessions. These tests of the technical functionalities, assessed qualitatively, were mainly to verify the correct working of the robotic systems.

For feedback from the volunteers, a questionnaire was designed, and the volunteer was asked to complete and answer the questions using a scale of 0 to 4 after being scanned by the robot. For the given score, 0 represents strongly disagree, 1 represents disagree, 2 represents neutral, 3 represents agree and 4 represents strongly agree. The questions relating to the use of robots are:

- Q1: I felt relaxed about the scan;
- Q2: The scanning robot appeared to be like a typical piece of hospital equipment;
- Q3: I found the appearance of the scanning robot to be appealing;
- Q4: I felt no discomfort during the scan;
- Q5: I felt no pain during the scan;
- Q6: I felt safe during the scan;
- Q7: I enjoyed the scanning experience.

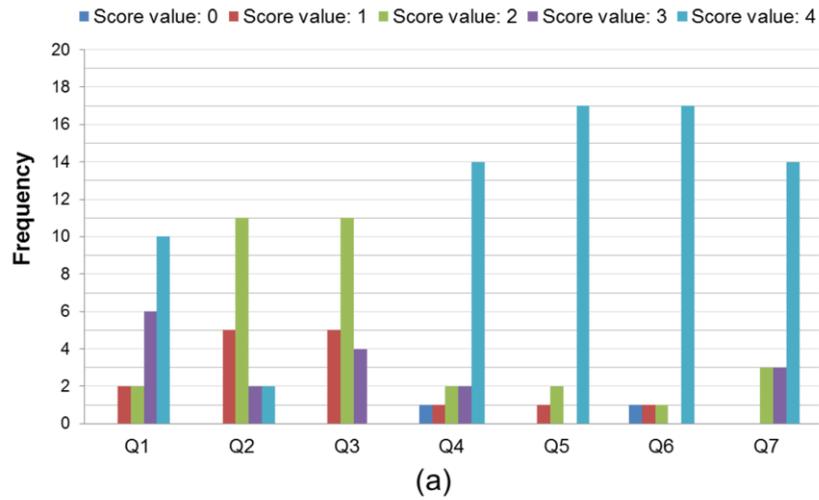

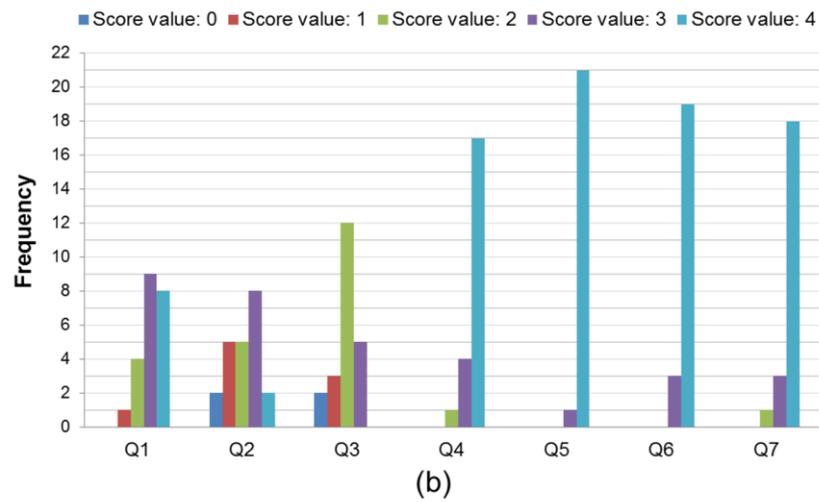

**Fig. 6.** Questionnaire results for the robotic ultrasound volunteer tests performed using (a) the iFIND-v2 robot (N=20) and (b) the iFIND-v3 robot (N =22).

The summary of the results of the questionnaire for the iFIND-v2 and iFIND-v3 robots (20 and 22 participants respectively) are shown in Fig 6. Most volunteers had positive experiences with the scan except with the appearance of the robots. They were neutral about the attractiveness and their similarity in appearance to hospital equipment. Comparing the two robots, a larger variation has been identified in terms of the similarity in appearance to hospital equipment for the iFIND-v3 dual-arm robot. Importantly, for both robots volunteers felt safe and reported little discomfort or pain while more consistent results have been identified for the iFIND-v3 dual-arm robot. However, there were outliers who did report discomfort, pain, and feeling unsafe for both robots and it could be useful to address this in any later designs.

For the iFIND-v2 robot only, we analysed the images obtainable, compared to the sonographer scanning manually. In each volunteer, the sonographer aimed to capture standard views of the aorta, including the following: pancreas transverse section (TS), left lobe of the liver TS, right lobe of the liver TS, right lobe of liver with right kidney, gallbladder longitudinal section, aorta at coeliac axis and aorta at mid abdominal position. Similar views were then targeted using the robot. The images were then scored by a sonographer for image quality as 'good', 'acceptable' and 'poor' according to the image quality component of the British Medical Ultrasound Society Peer Review Audit Tool 2014 v3 [17].

In total, 252 images were captured, 162 by sonographer and 90 by robot. Images from the first two volunteers were unlabelled and thus excluded from the analysis. The proportion of images with 'good' or 'acceptable' quality was 97.5% for sonographer and 81.1% for the robot, which is a statistically significant difference ($p<0.001$). Of the images with 'good' or 'acceptable' quality scores, the sonographer achieved a 'good' image in 72.2% of images, while the robot achieved this in 42.5% of images – again a statistically significant difference ($p<0.001$). When analysed by location, the robot was most often able to acquire the liver, pancreas and abdominal aorta images, which require a central or upper scanning area on the abdomen. It was often unable to acquire images of the gallbladder, kidneys, bladder and spleen, which require probe positions either on the side or caudal end of the abdomen.

Comparing only images of the liver, pancreas and abdominal aorta, which the robot was better able to capture, the sonographer achieved a 'good' or 'acceptable' quality in 96.6% of images, whereas the robot achieved this in 90.8% of images. This was not a statistically significant difference ($p>0.05$), which suggests the robot is capable holding the probe in contact with the necessary pressure to acquire adequate ultrasound images. However, the sonographer achieved significantly more 'good' images than 'acceptable' compared to the robot (74.4% compared to 52.4% respectively). This may be because of the more difficult indirect control when using the robot, making it harder to achieve the optimum image.

## 4 Discussion and Conclusions

With the general aim of using robotic technology to assist fetal ultrasound screening, we have developed three versions of the iFIND robots starting with a proof-of-concept prototype, coming to a significantly improved design with a patient-friendly appearance, and eventually finishing at a novel dual-arm robot for simultaneously controlling two ultrasound probes. With much more flexibility to make specially-shaped links and custom joint mechanisms, the iFIND robots look different from many existing robotic arms. The feedback from clinicians and patients indicated that these bespoke links can have positive impacts on the acceptance of using the robot in medicine. In terms of the robot configurations, we encountered great difficulties in finding the best arrangement of the DOFs in the design process and then solving the closed-form kinematics of the resulting configuration. Especially for the iFIND-v3 robot, the collision avoidance of the two arms and the various required arrangements of the two probes have led to a complicated kinematic analysis, which will be presented more technically separately.

In this design evolution process, apart from the technical considerations, one important driving factor was the feedback from the clinicians and patients. We realized that sometimes this is easily left out in the design process where the engineering team builds a robot to its technical expectation but the robot does not meet the expectation of the clinicians and patients in other aspects. Therefore, the iFIND robots were designed in a way that involved the combined inputs from engineers, clinicians, and patients. A number of examples can be found in our designs which were motivated by the clinicians and patients. These include the design of the end-effector unit for the iFIND-v2 robot, where we produced a streamlined 2 kg small unit incorporating five DOFs within 25 cm to improve the patients' and clinicians' acceptance. Similarly, the selection of the configuration of the gantry for the iFIND-v3 robot was based on the patients' inputs that they do not want to be enclosed while being scanned. We can conclude that these inputs are of great importance to our robot design.

In terms of the functionality and testing of the robot, a significant step for this project was to perform healthy volunteer tests in the design process and collect feedback from the volunteers. We realized that the only way to build confidence in using the robots for both engineers and clinicians is to continuously perform live tests. Looking at the results from the questionnaire, it is unsurprised to find out that the robots' appearances still need to be improved cosmetically to be like a piece of hospital equipment, although this is not the primary focus of this project. More importantly, the rest of the questions about comfort of the robots and their psychological effect indicate a good acceptability to be scanned by our robots for ultrasound examination. This is an important proof of our design idea.

When comparing image quality to images acquired by a sonographer, the iFIND-v2 robot was able to achieve a similar proportion of good or acceptable quality images in areas of the abdomen that it could easily reach. The unobtainable images are a limitation of the robot's workspace, which was designed for pregnant patients rather than non-pregnant volunteers. Therefore, the ability to reliably obtain some of the abdominal views is encouraging for the robots' abilities to scan a fetus in a pregnant patient. Currently the image qualities obtained do not reach the highly optimized quality achieved manually by a sonographer. However, this could improve with the development of a more sophisticated user interface, or perhaps automated optimization of the images using image quality feedback. It should also be noticed that the image quality study reported in this paper is still in an early stage while more systematic analyses with improved functionalities of the robot will be followed up for both robots in the future. With the current stage of the robots, it is difficult to compare the performance of the iFIND robots with the existing other robots in terms of the image acquisition quality as very limited clinical evidences in the literature are available for fully-active ultrasound robots used for abdominal scan.

From the technical point of view, we identified that the use of a custom-made mechanical clutch, with ball-spring pairs as the connection method between driven mechanism and the next link structure, is extremely useful. It not only prevents the joint from generating excessive force as a safety control independent of electrical systems and software logic, but also allows the operators and the volunteers to manually rotate each joint and move the robotic arm to other places, which turned out to be very useful in the real clinical scenario.

Working towards the future, we have developed a quality management system to facilitate the documenting and clinical translation of the robots and the goal is to eventually use the iFIND robots on pregnant women as the project is progressing. Importantly combining with the newly-developed image processing methods within the iFIND project, we intend to explore new ways of robotic-assisted ultrasound examination, which includes using the iFIND-v3 dual-arm robots to perform a full sweep of the abdominal area and extract useful information afterwards, compound the two ultrasound images from the two probes in real time to improve the visualization, and automatically detect the region and standard planes of the fetus using advanced machine learning algorithms and feedback to the robot for automatic adjustments.

**Acknowledgements.** This work was supported by the Wellcome Trust IEH Award [102431] and by the Wellcome/EPSRC Centre for Medical Engineering [WT203148/Z/16/Z]. The authors acknowledge financial support from the Department of Health via the National Institute for Health Research (NIHR) comprehensive Biomedical Research Centre award to Guy's & St Thomas' NHS Foundation Trust in partnership with King's College London and King's College Hospital NHS Foundation Trust.

## References

1. Priester, A.M., Natarajan, S., Culjat, M.O.: Robotic ultrasound systems in medicine. Ultrasonics, Ferroelectrics, and Frequency Control, IEEE Transactions on 60, 507-523 (2013)
2. Magnavita, N., Bevilacqua, L., Mirk, P., Fileni, A., Castellino, N.: Work-related musculoskeletal complaints in sonologists. Journal of Occupational and Environmental Medicine 41, 981-988 (1999)
3. LaGrone, L.N., Sadasivam, V., Kushner, A.L., Groen, R.S.: A review of training opportunities for ultrasonography in low and middle income countries. Tropical Medicine & International Health 17, 808-819 (2012)
4. Kilner, H., Wong, M., Walayat, M.: The antenatal detection rate of major congenital heart disease in Scotland. Scottish medical journal 56, 122-124 (2011)

5. Quartermain, M.D., Pasquali, S.K., Hill, K.D., Goldberg, D.J., Huhta, J.C., Jacobs, J.P., Jacobs, M.L., Kim, S., Ungerleider, R.M.: Variation in prenatal diagnosis of congenital heart disease in infants. Pediatrics peds. 2014-3783 (2015)
6. Vilchis Gonzales, A., Cinquin, P., Troccaz, J., Guerraz, A., Hennion, B., Pellissier, F., Thorel, P., Courreges, F., Gourdon, A., Poisson, G.: TER: a system for robotic tele-echography. In: Medical Image Computing and Computer-Assisted Intervention–MICCAI 2001, pp. 326-334. Springer, (2001)
7. Vilchis, A., Troccaz, J., Cinquin, P., Masuda, K., Pellissier, F.: A new robot architecture for tele-echography. Robotics and Automation, IEEE Transactions on 19, 922-926 (2003)
8. Masuda, K., Kimura, E., Tateishi, N., Ishihara, K.: Three dimensional motion mechanism of ultrasound probe and its application for tele-echography system. In: Intelligent Robots and Systems, 2001. Proceedings. 2001 IEEE/RSJ International Conference on, pp. 1112-1116. IEEE, (2001)
9. Gourdon, A., Poignet, P., Poisson, G., Vieyres, P., Marche, P.: A new robotic mechanism for medical application. In: Advanced Intelligent Mechatronics, 1999. Proceedings. 1999 IEEE/ASME International Conference on, pp. 33-38. IEEE, (1999)
10. Arbeille, P., Poisson, G., Vieyres, P., Ayoub, J., Porcher, M., Boulay, J.L.: Echographic examination in isolated sites controlled from an expert center using a 2-D echograph guided by a teleoperated robotic arm. Ultrasound in medicine & biology 29, 993-1000 (2003)
11. Arbeille, P., Ruiz, J., Herve, P., Chevillot, M., Poisson, G., Perrotin, F.: Fetal tele-echography using a robotic arm and a satellite link. Ultrasound in obstetrics & gynecology 26, 221-226 (2005)
12. Arbeille, P., Capri, A., Ayoub, J., Kieffer, V., Georgescu, M., Poisson, G.: Use of a robotic arm to perform remote abdominal telesonography. American journal of Roentgenology 188, W317-W322 (2007)
13. Courreges, F., Vieyres, P., Istepanian, R.: Advances in robotic tele-echography services-the OTELO system. In: Engineering in Medicine and Biology Society, 2004. IEMBS'04. 26th Annual International Conference of the IEEE, pp. 5371-5374. IEEE, (2004)
14. Vieyres, P., Poisson, G., Courrèges, F., Smith-Guerin, N., Novales, C., Arbeille, P.: A tele-operated robotic system for mobile tele-echography: The OTELO project.  M-Health, pp. 461-473. Springer (2006)
15. Noh, Y., Bimbo, J., Sareh, S., Wurdemann, H., Fraś, J., Chathuranga, D.S., Liu, H., Housden, J., Althoefer, K., Rhode, K.: Multi-Axis force/torque sensor based on Simply-Supported beam and optoelectronics. Sensors 16, 1936 (2016)
16. Wang, S., Housden, J., Noh, Y., Singh, A., Back, J., Lindenroth, L., Liu, H., Hajnal, J., Althoefer, K., Singh, D., Rhode, K.: Design and Implementation of a Bespoke Robotic Manipulator for Extra-corporeal Ultrasound. JoVE e58811 (2019)
17. https://www.bmus.org/static/uploads/resources/Peer_Review_Audit_Tool_wFYQwtA.pdf